\documentclass[letterpaper,journal,twoside]{IEEEtran}
\usepackage{amsmath,amsfonts}
\usepackage{algorithmic}
\usepackage{algorithm}
\usepackage{array}
\usepackage[caption=false,font=normalsize,labelfont=sf,textfont=sf]{subfig}
\usepackage{textcomp}
\usepackage{stfloats}
\usepackage{url}
\usepackage{verbatim}
\usepackage{graphicx}
\usepackage{cite}
\usepackage{booktabs}
\usepackage{multirow}
\usepackage{hyperref}

\begin{document}

\title{CReF: Cross-modal and Recurrent Fusion for Depth-conditioned Humanoid Locomotion}


\author{Yuan Hao$^{1}$, Ruiqi Yu$^{1}$, Shixin Luo$^{1}$, Guoteng Zhang$^{2}$, Jun Wu$^{1}$ and Qiuguo Zhu$^{*1}$%
\thanks{Manuscript received: April 22, 2026; Revised June 13, 2026; Accepted July 26, 2026.}
\thanks{This paper was recommended for publication by Editor Olivier Stasse upon evaluation of the Associate Editor and Reviewers comments. This work was supported by the ``Leading Goose'' R\&D Program of Zhejiang (Grant No. 2023C01177), the National Key R\&D Program of China (Grant No. 2022YFB4701502), and the 2035 Key Technological Innovation Program of Ningbo City (Grant No. 2024Z300).}%
\thanks{$^{1}$The authors are with Institute of Cyber-Systems and Control, Zhejiang University, Hangzhou 310027, China.}%
\thanks{$^{2}$Guoteng Zhang is with the School of Control Science and Engineering, Shandong University, Jinan 250061, China.}%
\thanks{$^{*}$Corresponding author: Qiuguo Zhu (\texttt{qgzhu@zju.edu.cn}).}
\thanks{Project website: \protect\url{https://cometlogic.github.io/cref/}.}%
\thanks{Digital Object Identifier (DOI): see top of this page.}%
}

\markboth{IEEE ROBOTICS AND AUTOMATION LETTERS. PREPRINT VERSION. ACCEPTED JULY, 2026}%
{Hao \MakeLowercase{\textit{et al.}}: CReF: Cross-modal and Recurrent Fusion for Depth-conditioned Humanoid Locomotion}


\maketitle

\begin{abstract}

Stable traversal over geometrically complex terrain increasingly requires exteroceptive perception, yet prior perceptive humanoid locomotion methods often remain tied to explicit geometric abstractions, either by mediating control through robot-centric 2.5D terrain representations or by shaping depth learning with auxiliary geometry-related targets. While effective, these approaches introduce additional map-construction procedures or multi-stage skill-transfer processes beyond direct depth-to-control learning. We propose CReF (Cross-modal and Recurrent Fusion), a single-stage depth-conditioned humanoid locomotion framework that learns locomotion-relevant features directly from raw forward-facing depth without explicit geometric intermediates. CReF couples proprioception and depth tokens through proprioception-queried cross-modal attention, fuses the resulting representation with a gated residual fusion block, and performs temporal integration with a Gated Recurrent Unit (GRU) regulated by a highway-style output gate for state-dependent blending of recurrent and feedforward features. To further improve terrain interaction, we introduce a terrain-aware foothold placement reward that extracts supportable foothold candidates from foot-end point-cloud samples and rewards touchdown locations that lie close to the nearest supportable candidate. Experiments in simulation and on a physical humanoid demonstrate robust traversal over diverse terrains and effective zero-shot transfer to real-world scenes containing handrails, hollow pallet assemblies, severe reflective interference, and visually cluttered outdoor surroundings.

\end{abstract}


\section{Introduction}
\label{sec:intro}
\IEEEPARstart{H}{umanoid} robots promise versatile mobility in human-centered environments. Deep reinforcement learning has enabled agile locomotion across diverse platforms and surface conditions, and proprioception-dominant policies can already achieve strong standing and walking performance through reactive stabilization \cite{hwangbo2019agile, choi2023deformable, van2024revisiting, zhang2024learning}. However, perceptive locomotion becomes essential when safe traversal depends on anticipating terrain geometry before contact. This is especially critical on stairs, particularly during descent, and near gaps or abrupt height transitions, where foothold errors can lead to edge contacts or slipping \cite{Long2025PIM, song2025gait}.

\begin{figure}[t]
  \centering
  \includegraphics[width=\linewidth]{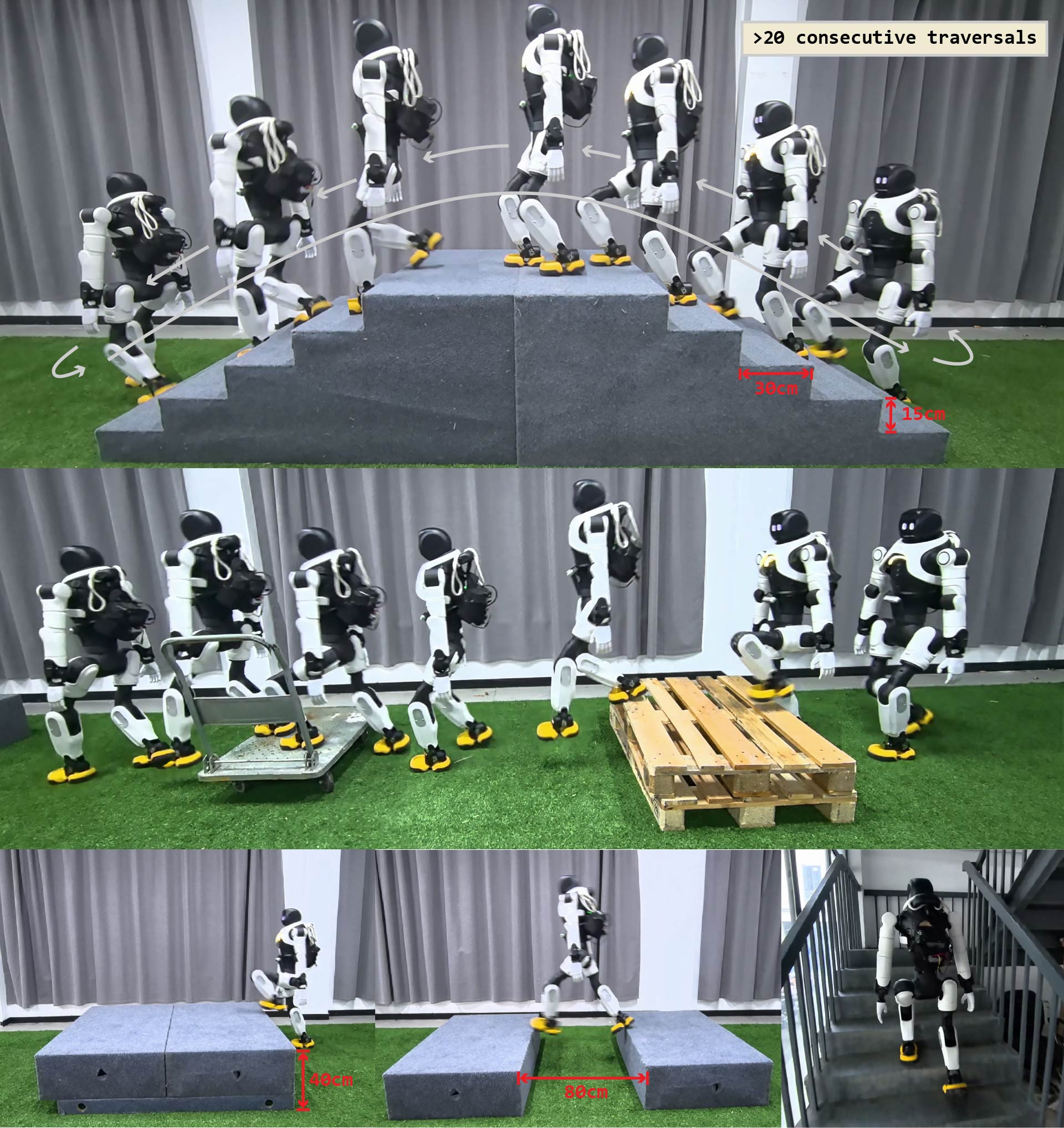}
  \caption{Our CReF framework enables robust real-world humanoid locomotion, including more than 20 consecutive stair traversals, a 40\,cm high platform, an 80\,cm gap, and real-world stairs with a 20\,cm rise and 26\,cm tread, while generalizing robustly beyond the training terrains.}
  \label{fig:cover}
\end{figure}

A widely adopted way to incorporate exteroception into legged locomotion is to construct a robot-centric 2.5D terrain representation for foothold selection, predictive control, or policy conditioning \cite{jenelten2020perceptive,grandia2023perceptive,miki2022learning,wang2025beamdojo,dai2026walk}. Hybrid planner-guided methods and attention-based map encoders further extend this paradigm by providing structured references or emphasizing future steppable regions \cite{jenelten2024dtc,he2025attention}. This line of work is attractive because elevation maps offer a compact, interpretable perception-control interface. However, point-cloud data must first be projected, fused, filtered, and maintained in a consistent frame before being used for downstream planning and control. This process often involves smoothing, inpainting, segmentation, terrain classification, or drift compensation. \cite{miki2022elevation,grandia2023perceptive}. Forward-facing depth provides an egocentric geometric signal directly from the onboard camera, reducing reliance on a separate terrain-map construction pipeline.

A further line of work uses depth as the exteroceptive input, but often shapes depth learning with auxiliary targets \cite{cheng2024extreme,liu2026faststairlearningrunstairs} because forward facing depth provides look ahead cues without directly observing the instantaneous underfoot region \cite{yang2023neural, zhuang2023robot}.
Rather than letting the locomotion objective alone organize the visual representation, these methods regularize the depth branch through terrain reconstruction \cite{yu2025start,song2025gait,duan2024learning}, decoder-constrained latent estimation \cite{luo2024pie}, or transfer from privileged teachers \cite{zhuang2024humanoid}. This strategy is attractive because geometry-related targets provide structured guidance that can ease optimization and improve transfer. However, it also changes how depth features are learned. The encoder is encouraged to preserve information supporting the chosen auxiliary target, making the representation partly dependent on the target design and fidelity rather than solely on the locomotion objective. A more direct and simpler formulation is to learn locomotion-relevant structure from depth without explicit terrain reconstruction or privileged geometric targets. Recent humanoid work has begun to pursue this route by training end-to-end policies that consume raw depth and proprioception directly, using temporal depth history to mitigate partial observability and improve traversal over challenging terrain and obstacle sequences \cite{zhu2026hiking}. In parallel, recent perceptive humanoid pipelines have reduced the depth sensing sim-to-real gap through transfer-oriented training and realistic sensor simulation \cite{zhu2026hiking,sun2026nowyouseethat}. These methods are effective, but simulated artifacts may not fully capture real sensor failures. We therefore examine whether a depth-conditioned policy can maintain stable locomotion under real sensing artifacts without precise depth-noise modeling during training.

In this work, we propose CReF, a perceptive humanoid locomotion policy that directly maps proprioception and raw forward-facing depth to joint position targets without explicit geometric intermediates or multi-stage supervision. This design explores whether a single-stage policy can learn locomotion-relevant structure directly from raw depth and achieve or even exceed the performance of geometry-guided methods. CReF combines cross-modal attention, gated residual fusion, and recurrent fusion to extract depth features and modulate temporal memory based on the locomotion state. We further introduce a terrain-aware foothold placement reward that extracts supportable foothold candidates from local foot-end point-cloud samples and shapes touchdown toward them. Unlike purely prohibitive do-not-step constraints \cite{wang2025beamdojo,zhuang2024humanoid}, it encourages landing near supportable regions and thus provides more directional supervision for contact placement. CReF transfers zero-shot to real hardware and remains effective in challenging real-world scenes. 

This work makes the following contributions:
\begin{itemize}
    \item A single-stage depth-conditioned humanoid locomotion framework that organizes raw forward-facing depth and proprioception into locomotion-relevant multimodal and temporal representations, enabling robust terrain traversal without explicit geometric intermediates or multi-stage supervision.
    \item A terrain-aware foothold placement reward that extracts supportable foothold candidates from local foot-end point-cloud samples and guides touchdown during terrain transitions, yielding substantial gains in long-duration stair traversal, especially in descent.
    \item Real-world zero-shot deployment of a single depth-conditioned humanoid locomotion policy across multiple terrain-transition tasks and broader unseen environments, demonstrating robust transfer beyond simulation.
\end{itemize}

\section{Method}
\label{sec:method}

\subsection{Overview}
\label{subsec:method_overview}

\begin{figure*}[t]
  \centering
  \vspace{7pt}
  \includegraphics[width=1.0\textwidth]{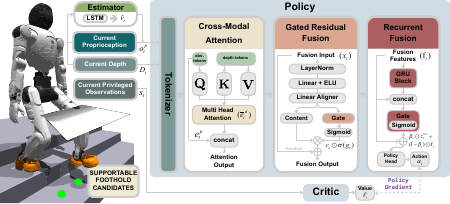}
  \caption{Overview of CReF. The proposed single-stage depth-conditioned policy combines cross-modal attention, gated residual fusion, recurrent fusion, and a terrain-aware foothold placement reward for robust terrain locomotion.}
  \label{fig:cref_arch}
\end{figure*}

CReF is a single-stage depth-conditioned humanoid locomotion framework that maps onboard proprioception and forward-facing depth directly to joint position targets. It comprises an onboard actor, an asymmetric critic with privileged training signals, and a lightweight velocity estimator. The architecture is shown in Fig.~\ref{fig:cref_arch}, and training is performed with PPO \cite{schulman2017ppo}.

\subsubsection{Policy Network}
At each control step $t$, the policy network takes the current proprioceptive observation $\mathbf{o}^{p}_t$ and the current forward-facing depth image $\mathbf{D}_t\in\mathbb{R}^{H\times W}$ as input:
\begin{equation}
\mathbf{o}_t = \big(\mathbf{o}^{p}_t,\; \mathbf{D}_t\big).
\label{eq:actor_obs}
\end{equation}
Following standard legged locomotion pipelines, the proprioceptive observation is defined as
\begin{equation}
\mathbf{o}^{p}_t =
\Big[
\boldsymbol{\omega}_t,\;
\mathbf{r}^{\mathrm{grav}}_t,\;
\mathbf{u}^{\mathrm{cmd}}_t,\;
\mathbf{q}_t-\mathbf{q}_0,\;
\dot{\mathbf{q}}_t,\;
\mathbf{a}_{t-1}
\Big],
\label{eq:proprio}
\end{equation}
where $\boldsymbol{\omega}_t$ is the base angular velocity, $\mathbf{r}^{\mathrm{grav}}_t$ is the gravity direction expressed in the body frame, $\mathbf{u}^{\mathrm{cmd}}_t$ is the commanded motion, $\mathbf{q}_t$ and $\dot{\mathbf{q}}_t$ are the joint positions and velocities, $\mathbf{q}_0$ is the nominal standing pose, and $\mathbf{a}_{t-1}$ is the previous action.

\subsubsection{Value Network}
To obtain a more accurate estimate of the state value, the value network is trained asymmetrically with privileged information available only in simulation. In addition to the proprioceptive observation $\mathbf{o}^{p}_t$, its input includes the ground-truth base linear velocity $\mathbf{v}_t$ and a robot-centric terrain height observation $\mathbf{m}_t$, which is defined as local elevation samples around the robot. The value-network input is
\begin{equation}
\mathbf{s}_t = \big[\mathbf{o}^{p}_t,\; \mathbf{v}_t,\; \mathbf{m}_t\big].
\label{eq:critic_state}
\end{equation}

\subsubsection{Action space}
The policy outputs an action $\mathbf{a}_t\in\mathbb{R}^{n_a}$ that parameterizes joint position targets tracked by a low-level PD controller. We adopt a residual parameterization around the nominal standing pose $\mathbf{q}_0$:
\begin{equation}
\mathbf{q}^{\mathrm{target}}_t = \mathbf{q}_0 + \mathbf{a}_{t},
\label{eq:action}
\end{equation}
where $\mathbf{q}_0$ is the nominal standing pose and $\mathbf{q}^{\mathrm{target}}_t$ is the commanded joint position vector.

\subsubsection{Reward function}

Following recent analyses on robust humanoid reward design \cite{rudin2022learning,van2024revisiting}, we adopt a low-speed-aware linear-velocity tracking term and a contact-shaping term that rewards standing while encouraging short-horizon single-foot contact during locomotion. The reward terms used in training are listed in Table~\ref{tab:rewards}.

\begin{table}[t]
    \caption{REWARD TERMS USED IN TRAINING.}
    \label{tab:rewards}
    \centering
    \scriptsize
    \setlength{\tabcolsep}{2.5pt}
    \renewcommand{\arraystretch}{1.08}
    \begin{tabular}{p{2.05cm} p{4.4cm} c}
        \toprule
        Term & Expression ($r_i$) & $w_i$ \\
        \midrule
        Lin. vel. tracking &
        $\exp\!\left(-\frac{\mathbb{I}_s\|\mathbf{e}_{v}\|_1+(1-\mathbb{I}_s)\|\mathbf{e}_{v}\|_2^2}{\sigma_v}\right)$
        & $2.5$ \\
        Ang. vel. tracking &
        $\exp\!\left(-(\omega_{z}-\omega^{\mathrm{cmd}}_{z})^2/\sigma_\omega\right)$
        & $1.5$ \\
        Orientation &
        $\exp\!\left(-10\|\mathbf{r}^{\mathrm{grav}}_{xy}\|_2\right)$
        & $1.0$ \\
        Feet contact &
        $\mathbb{I}_{\mathrm{stand}}+(1-\mathbb{I}_{\mathrm{stand}})\,\mathbb{I}^{0.2\mathrm{s}}_{\mathrm{single}}$
        & $1.5$ \\
        Feet $y$ distance &
        $\exp\!\left(-10|d^{\mathrm{feet}}_y-0.27|\right)$
        & $0.1$ \\
        Foothold placement &
        Eq.~\eqref{eq:foothold_reward}
        & $2.0$ \\
        Base height &
        $\exp\!\left(-10|(z^{\mathrm{base}}-\bar{z}^{\mathrm{supp}})-0.55|\right)$
        & $0.8$ \\
        \midrule
        Lin. vel. $z$ &
        $v_z^2$
        & $-1.0$ \\
        Ang. vel. $xy$ &
        $\|\boldsymbol{\omega}_{xy}\|^2$
        & $-0.05$ \\
        Joint velocity &
        $\|\dot{\mathbf{q}}\|^2$
        & $-10^{-3}$ \\
        Action rate &
        $\|\mathbf{a}_t-\mathbf{a}_{t-1}\|^2$
        & $-0.01$ \\
        Action smoothness &
        $\|\mathbf{a}_t-2\mathbf{a}_{t-1}+\mathbf{a}_{t-2}\|^2$
        & $-0.01$ \\
        Feet air time &
        $\sum_{f}\mathbb{I}^{f}_{\mathrm{first}}\left(0.5-T^{f}_{\mathrm{air}}\right)$
        & $-2.5$ \\
        Foot slip &
        $\sum_f \mathbb{I}\!\left(\|\mathbf{F}^f\|>5\right)\,\|\mathbf{v}^{f}_{xyz}\|$
        & $-0.2$ \\
        Foot impact acc. &
        $\sum_f \big[\,|\dot{v}^{f}_{z}|-a_0\,\big]_+$
        & $-0.5$ \\
        Foot impact vel. &
        $\sum_f \big[\,|v^{f}_{z}|-v_0\,\big]_+$
        & $-1.5$ \\
        DoF position limits &
        $\sum_j\!\left(\big[q^{\min}_j-q_j\big]_+ + \big[q_j-q^{\max}_j\big]_+\right)$
        & $-10$ \\
        DoF velocity limits &
        $\sum_j \big[\,|\dot{q}_j|-\dot{q}^{\max}_j\,\big]_+$
        & $-0.6$ \\
        Hip pos &
        $\sum_{j\in\mathcal{J}_{\mathrm{hip\ x/z}}} (q_j-q^{0}_j)^2$
        & $-7$ \\
        Ankle pos &
        $\sum_{j\in\mathcal{J}_{\mathrm{ank\ x}}} (q_j-q^{0}_j)^2$
        & $-10$ \\
        Stumble &
        $\mathbb{I}\!\left(\exists f:\|\mathbf{F}^{f}_{xy}\| > 5\,|F^{f}_{z}|\right)$
        & $-10$ \\
        \bottomrule
    \end{tabular}
\end{table}

Here, $\mathbf{e}_v=\mathbf{u}^{\mathrm{cmd}}_{xy}-\mathbf{v}_{xy}$ is the planar velocity tracking error, and $\mathbb{I}_s=\mathbb{I}(\|\mathbf{u}^{\mathrm{cmd}}_{xy}\|_2<v_s)$ switches to the low-speed form when the commanded planar speed is below $v_s$. The standing indicator is defined as $\mathbb{I}_{\mathrm{stand}}=\mathbb{I}(\|\mathbf{u}^{\mathrm{cmd}}\|_2<u_s)$. $\mathbb{I}^{0.2\mathrm{s}}_{\mathrm{single}}$ is equal to 1 if a single-foot contact has occurred within the last 0.2\,s, and 0 otherwise. We use $\mathbf{r}^{\mathrm{grav}}_{xy}$ for the $xy$ component of the projected gravity direction, and
\begin{equation}
\bar{z}^{\mathrm{supp}}
=
\frac{\sum_f \mathbb{I}(F^f_z>1)\, z^f}
     {\sum_f \mathbb{I}(F^f_z>1)}
\end{equation}
for the average height of supporting feet.

In Table~\ref{tab:rewards}, the superscript $f$ indexes feet. Accordingly, $\mathbf{F}^f$ and $\mathbf{v}^f$ denote the contact force and velocity of foot $f$, while $F^f_z$ and $\mathbf{F}^f_{xy}$ denote its normal and tangential contact-force components. $\mathbb{I}^{f}_{\mathrm{first}}$ indicates the first contact event of foot $f$ after an aerial phase, and $T^f_{\mathrm{air}}$ is the corresponding air time. We use $[x]_+=\max(x,0)$ for the positive-part operator, and $q_j^0$ denotes the $j$-th component of the standing pose $\mathbf{q}_0$.

\subsection{Policy Architecture}
\label{subsec:method_arch}

The policy is designed to extract state-relevant terrain features from forward-facing depth observations and to adaptively control the contribution of temporal memory according to the current locomotion state. In what follows, we refer to the fusion block after cross-modal attention as gated residual fusion (GRF), and to the temporal module as recurrent fusion, which consists of GRU-based temporal integration followed by a highway output gate. The architectural components are detailed in the following subsections.

\subsubsection{Depth encoding and velocity estimation}
\label{subsubsec:depth_vel}

The raw depth image is normalized as $\mathbf{D}_t=\mathbf{D}^{\mathrm{raw}}_t / d_{\max}-0.5$, where $d_{\max}$ is the preset maximum depth range. Then the normalized depth image is encoded by a lightweight CNN-based depth tokenizer $\mathcal{T}_{\theta}(\cdot)$ into a set of local depth tokens,
\begin{equation}
\mathbf{Z}_t = \mathcal{T}_{\theta}(\mathbf{D}_t)\in\mathbb{R}^{N\times d},
\label{eq:depth_token}
\end{equation}
where $N$ is the number of tokens and $d$ is the token dimension. These depth tokens are used by the policy for subsequent cross-modal attention.

To facilitate velocity-command tracking, we jointly train an auxiliary base-velocity estimator,
\begin{equation}
\hat{\mathbf{v}}_t
=
\mathrm{LSTM}\!\left([\mathbf{o}^{p}_t;\mathcal{C}_{\psi}(\mathbf{D}_t)]\right),
\label{eq:vel_est}
\end{equation}
where $\mathcal{C}_{\psi}(\cdot)$ is a lightweight depth compression module. The estimator is supervised by the simulator ground-truth base linear velocity using an $\ell_2$ loss.

\subsubsection{Cross-modal attention for proprioception-conditioned depth feature extraction}
\label{subsubsec:attn}

The proprioceptive tokenizer takes the concatenation of proprioception and the estimated base linear velocity as input:
\begin{equation}
\mathbf{e}^p_t = \mathcal{P}_{\phi}\!\left([\mathbf{o}^{p}_t;\hat{\mathbf{v}}_t]\right), \qquad
\mathbf{E}^d_t = \mathrm{LN}(\mathbf{Z}_t),
\label{eq:obs_depth_embed}
\end{equation}
where $\mathcal{P}_{\phi}(\cdot)$ denotes the proprioceptive tokenizer and $\mathbf{E}^d_t$ is the normalized depth-token set.

Cross-modal attention then uses the proprioceptive token as the query and the depth tokens as the keys and values:
\begin{equation}
\mathbf{Q}_t = \mathrm{LN}(\mathbf{e}^p_t)\mathbf{W}_q,\qquad
\mathbf{K}_t = \mathbf{E}^d_t\mathbf{W}_k,\qquad
\mathbf{V}_t = \mathbf{E}^d_t\mathbf{W}_v,
\end{equation}
where $\mathbf{Q}_t\in\mathbb{R}^{1\times d}$ and $\mathbf{K}_t,\mathbf{V}_t\in\mathbb{R}^{N\times d}$. Multi-head attention (MHA) then aggregates locomotion-relevant information from the depth tokens conditioned on the current proprioceptive state. The resulting attention output is
\begin{equation}
\bar{\mathbf{e}}^d_t = \mathrm{MHA}\!\big(\mathbf{Q}_t,\mathbf{K}_t,\mathbf{V}_t\big),
\label{eq:attn_fuse}
\end{equation}
which is concatenated with the proprioceptive token to form the fusion input,
\begin{equation}
\mathbf{x}_t = [\mathbf{e}^p_t;\bar{\mathbf{e}}^d_t].
\end{equation}

\subsubsection{Gated residual fusion and recurrent fusion}
\label{subsubsec:gating}

Given the fusion input $\mathbf{x}_t$, the GRF first computes
\begin{equation}
\tilde{\mathbf{x}}_t
= \phi\!\big(\mathbf{W}_1\,\mathrm{LN}(\mathbf{x}_t)+\mathbf{b}_1\big),
\end{equation}
where $\phi(\cdot)$ is ELU. This projection mixes the proprioceptive and depth-conditioned features in a shared latent space while keeping the block lightweight.

A second linear layer then produces a content branch and a gate branch,
\begin{equation}
\begin{bmatrix}
\mathbf{c}_t\\
\mathbf{g}_t
\end{bmatrix}
=
\mathbf{W}_2\,\tilde{\mathbf{x}}_t+\mathbf{b}_2,
\qquad
\mathbf{c}_t,\mathbf{g}_t\in\mathbb{R}^{2d},
\end{equation}
and the fusion output is
\begin{equation}
\mathbf{f}_t
= \mathbf{x}_t + \mathbf{c}_t\odot\sigma(\mathbf{g}_t).
\label{eq:grn}
\end{equation}
Here, $\mathbf{c}_t$ provides a candidate residual update, while $\sigma(\mathbf{g}_t)$ adaptively controls its channel-wise contribution. This design allows the policy to emphasize informative multimodal corrections while preserving a direct residual path for stable optimization.

Within recurrent fusion, $\mathbf{f}_t$ is first processed by a GRU for temporal integration,
\begin{equation}
\mathbf{h}_t = \mathrm{GRU}(\mathbf{f}_t,\mathbf{h}_{t-1}),\qquad
\mathbf{z}^{\mathrm{rec}}_t = \mathbf{W}_h\mathbf{h}_t,
\end{equation}
where $\mathbf{z}^{\mathrm{rec}}_t\in\mathbb{R}^{2d}$ is the recurrent feature. The GRU aggregates short-horizon temporal context, which is important when a single depth frame is insufficient to disambiguate terrain structure or contact timing.

The recurrent feature is then combined with the current fused feature through a highway output gate,
\begin{equation}
\beta_t = \sigma\!\big(\mathbf{W}_{\beta}[\mathbf{z}^{\mathrm{rec}}_t;\mathbf{f}_t]+\mathbf{b}_{\beta}\big), \qquad
\mathbf{y}_t = \beta_t\odot\mathbf{z}^{\mathrm{rec}}_t + (1-\beta_t)\odot\mathbf{f}_t.
\label{eq:highway}
\end{equation}
This gate enables state-dependent fusion between instantaneous multimodal evidence and recurrent memory, so that the policy can rely more on temporal features in ambiguous or risky phases while retaining stronger feedforward responses when current observations are already sufficient. The final action is produced by an MLP head from $\mathbf{y}_t$.

\subsection{Terrain-Aware Foothold Placement Reward}
\label{subsec:method_foothold}

We introduce a terrain-aware foothold placement reward to shape foot placement at touchdown. For each foot, we maintain a compact local point buffer of $60\times5$ foot-frame samples. At each update, command-conditioned forward gating removes near-foot points, where the minimum retained forward distance is set by the commanded step extent and taken as the distance covered in 0.5\,s under the current forward command. The remaining points are partitioned into overlapping candidate windows of size $24\,cm \times 10\,cm$ with a stride of $4\,cm$. To reduce computation and chattering, candidates are updated only at liftoff and fixed until touchdown.

Let $\mathcal{P}^{f}_{t,k}=\{\mathbf{p}^{f}_{t,k,i}\}_{i=1}^{n^{f}_{t,k}}$ denote the $k$-th candidate window for foot $f$ at time $t$, where $n^{f}_{t,k}$ is the number of points in that window. For each window, we compute the mean and covariance,
\begin{equation}
\begin{aligned}
\boldsymbol{\mu}^{f}_{t,k} 
&= \frac{1}{n^{f}_{t,k}}
   \sum_{i=1}^{n^{f}_{t,k}} \mathbf{p}^{f}_{t,k,i}, \\
\mathbf{\Sigma}^{f}_{t,k} 
&= \frac{1}{\max(n^{f}_{t,k}-1,1)} \\
&\quad  
\sum_{i=1}^{n^{f}_{t,k}}
\big(\mathbf{p}^{f}_{t,k,i}-\boldsymbol{\mu}^{f}_{t,k}\big)
\big(\mathbf{p}^{f}_{t,k,i}-\boldsymbol{\mu}^{f}_{t,k}\big)^\top .
\end{aligned}
\label{eq:cov}
\end{equation}
We then perform eigen-decomposition,
\begin{equation}
\Sigma_{t,k}^f v_{t,k,j}^f = \lambda_{t,k,j}^f v_{t,k,j}^f,
\qquad
\lambda_{t,k,1}^f \le \lambda_{t,k,2}^f \le \lambda_{t,k,3}^f .
\end{equation}
and define the roughness as
\begin{equation}
\rho^{f}_{t,k}=\sqrt{\max(\lambda^{f}_{t,k,1},0)}.
\end{equation}
A window is accepted as a foothold candidate if it is sufficiently planar, approximately horizontal, and not recessed:
\begin{equation}
\rho_{t,k}^f < r_{\mathrm{th}},\qquad
|v_{t,k,1,z}^f| > \eta_{\mathrm{th}},\qquad
\mu_{t,k,z}^f > h_{\min},
\end{equation}
For each accepted window, the candidate foothold is set to its mean,
\begin{equation}
\mathbf{p}^{f,\star}_{t,k}=\boldsymbol{\mu}^{f}_{t,k}.
\end{equation}

To avoid chattering, the candidate set is refreshed only at liftoff and then held fixed until the next touchdown. At touchdown, we compute the nearest planar distance in the $x$-$z$ plane between the realized contact position and the candidate set,
\begin{equation}
d^{f}_{xz}
=
\min_k
\left\|
\begin{bmatrix}
p^{f}_{x,t}\\
p^{f}_{z,t}
\end{bmatrix}
-
\begin{bmatrix}
p^{f,\star}_{x,t,k}\\
p^{f,\star}_{z,t,k}
\end{bmatrix}
\right\|_2,
\end{equation}
and define the reward as
\begin{equation}
r_{\mathrm{fh}}
=
\sum_{f\in\mathcal{F}}
I^{f}_{\mathrm{td}}
\exp\!\left(-d^{f}_{xz}/s_{xz}\right),
\label{eq:foothold_reward}
\end{equation}
where $I^{f}_{\mathrm{td}}$ indicates a touchdown event and $s_{xz}$ is a tolerance scale. Although computed at touchdown, this term provides an anticipatory shaping objective by encouraging the swing foot to approach locally supportable regions before contact.

\subsection{Training Details}
\label{subsec:method_training}

CReF is trained in Isaac Gym with 4096 parallel environments at a control frequency of 50\,Hz. The depth stream is rendered at $64\times48$ resolution and updated at $20$~Hz. The terrain-aware foothold placement reward adds a small training overhead. It increases the wall clock time by approximately 0.5\,s per iteration. The full CReF policy still completes training within about 30 hours for 20{,}000 iterations on a single RTX~4090.


To support large-batch training, depth rendering is implemented with NVIDIA Warp \cite{warp2022}. Robot links are approximated as capsules, and self-occlusion is computed on the GPU via first-hit ray-capsule intersection. No synthetic depth corruption is injected during training.




\section{Experiments}
\label{sec:exp}

We evaluate CReF in simulation and on hardware. The simulation experiments examine three questions: (i) whether the proposed framework improves terrain traversal performance relative to ablated variants and a perceptive locomotion baseline, (ii) whether the foothold placement reward improves stair contact placement, and (iii) how the highway gate recruits recurrent features under different traversal conditions. We then assess zero-shot real-world transfer and out-of-distribution robustness on a physical humanoid robot.

\subsection{Experimental Setup}
\label{subsec:exp_setup}

Our deployment platform is an AGIBOT X2 Ultra humanoid \cite{x2}, which is approximately 1.31\,m tall, weighs 39\,kg. We choose Humanoid Parkour Learning (HPL)~\cite{zhuang2024humanoid} as the main external baseline because it is a representative teacher-student perceptive parkour framework that learns a deployable depth-based policy under guidance from privileged terrain information, providing a paradigm-level comparison to our single-stage raw-depth design. Since HPL was originally developed for the Unitree H1~\cite{h1}, we re-implement it on the AGIBOT X2 Ultra in simulation to ensure hardware-controlled comparison.

For onboard sensing and inference, the robot uses a forward-facing Intel RealSense D435i mounted with a downward pitch of $50^\circ$ relative to the horizontal plane, together with an NVIDIA Jetson AGX Orin for onboard policy inference. The policy is transferred zero-shot from simulation, without task-specific fine-tuning. Using the hardware timestamps provided by the RealSense, we fit a linear regression to map camera time to system time and select the depth frame with an effective latency of 20\,ms.

We compare the following policy variants, all trained from scratch under matched reward, optimization, and terrain settings:
\begin{itemize}
    \item \textbf{Full CReF}: cross-modal attention + gated residual fusion (GRF) + recurrent fusion.
    \item \textbf{w/o Cross-Attn}: removes cross-modal attention while preserving the remaining architecture.
    \item \textbf{w/o GRF}: replaces gated residual fusion with a residual MLP block of matched depth and width.
    \item \textbf{w/o Highway Gate}: removes the highway output gate from recurrent fusion and decodes actions directly from the GRU feature.
    \item \textbf{HPL}: a perceptive locomotion baseline trained and evaluated under the same terrain protocol.
\end{itemize}

\begin{figure}[t]
    \centering
    \vspace{7pt}
    \includegraphics[width=0.8\linewidth]{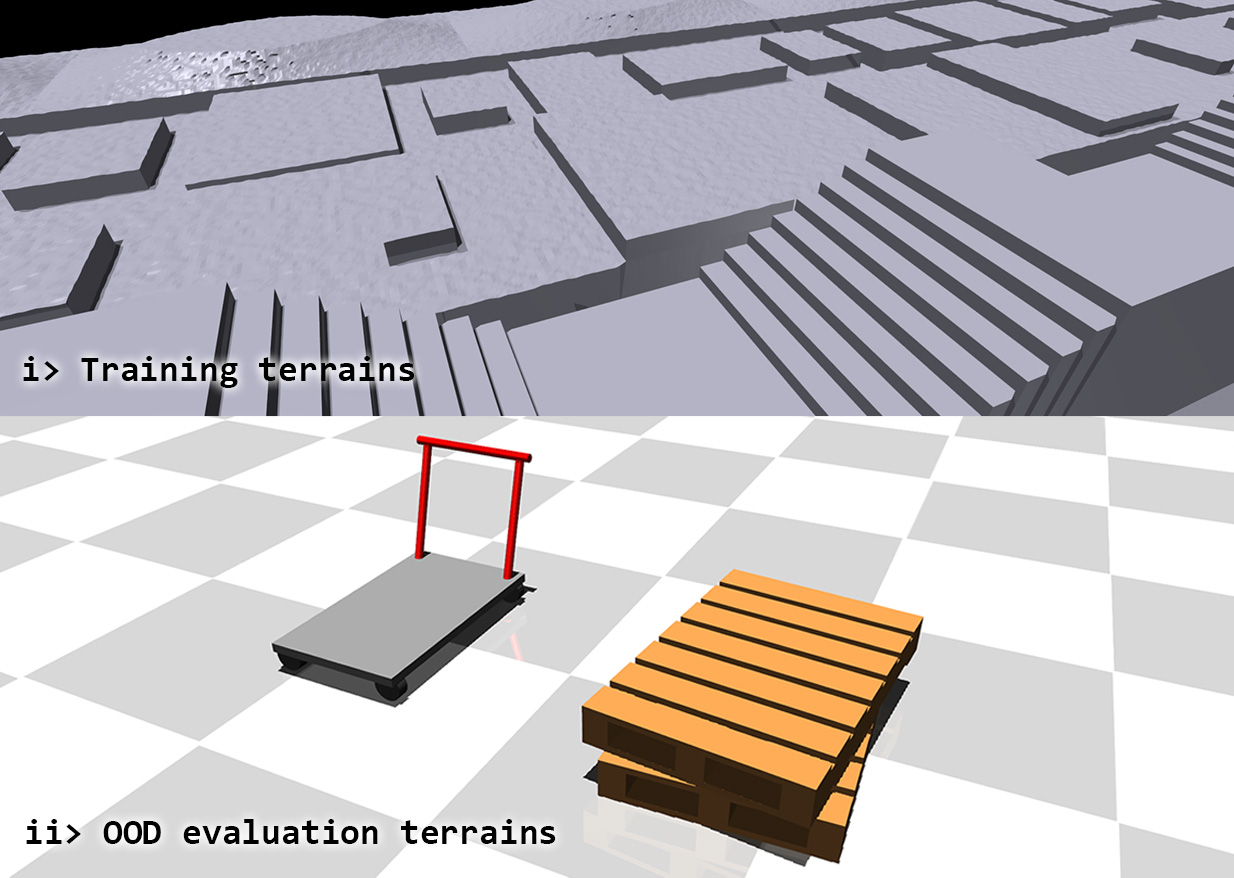}
    \caption{Simulation terrains used in the experiments. The upper panel shows representative terrains used during training. The lower panel presents additional MuJoCo out-of-distribution terrains for cross-simulator evaluation, including terrains with perforated or hollow foothold surfaces that induce depth-distribution shift through missing or discontinuous support geometry, as well as scenes containing lateral heterogeneous structures (e.g., trailer handrails) that introduce cluttered side geometry absent from training.}
    \label{fig:terrain_setup}
\end{figure}

For architectural comparison, each policy is evaluated in 2000 parallel environments for up to 40\,s, with forward commands randomly sampled from 0.4--0.8\,m/s. Failure is defined as becoming unable to recover to normal walking. For stair foot-placement analysis, we use 2048 parallel stair rollouts. For highway-gate analysis, we log gate statistics from 4096 parallel environments under a fixed forward command. MuJoCo \cite{todorov2012mujoco} is used only for an additional out-of-distribution evaluation in the architectural comparison.  Fig.~\ref{fig:terrain_setup} shows the simulated training terrains together with the out-of-distribution evaluation terrains.

For the foothold-objective comparison, we replace the foothold placement reward in Eq.~\eqref{eq:foothold_reward} with a foot contact quality reward (FCQR) following BeamDojo~\cite{wang2025beamdojo}, and also evaluate a no-foothold-reward variant (NoFR), while keeping the remaining training settings unchanged.

\begin{table*}[t]
    \centering
    \vspace{10pt}
    \caption{ARCHITECTURAL COMPARISON OF TERRAIN TRAVERSAL SUCCESS RATES (\%).}
    \label{tab:arch_comp}
    \scriptsize
    \setlength{\tabcolsep}{3.4pt}
    \renewcommand{\arraystretch}{1.10}
    \resizebox{\textwidth}{!}{
    \begin{tabular}{lcccccccccccccc}
        \toprule
        \multirow{2}{*}{Method}
        & \multicolumn{4}{c}{Stairs, SR}
        & \multicolumn{4}{c}{Gap, SR}
        & \multicolumn{4}{c}{Platform, SR}
        & \multirow{2}{*}{\begin{tabular}[c]{@{}c@{}}MuJoCo\\OOD\end{tabular}}
        & \multirow{2}{*}{\begin{tabular}[c]{@{}c@{}}Overall\\(w/o MuJoCo)\end{tabular}} \\
        \cmidrule(lr){2-5}\cmidrule(lr){6-9}\cmidrule(lr){10-13}
        & \begin{tabular}[c]{@{}c@{}}Easy\\(10/30 cm)\end{tabular}
        & \begin{tabular}[c]{@{}c@{}}Medium\\(15/30 cm)\end{tabular}
        & \begin{tabular}[c]{@{}c@{}}Hard\\(20/30 cm)\end{tabular}
        & \begin{tabular}[c]{@{}c@{}}OOD\\(23/30 cm)\end{tabular}
        & \begin{tabular}[c]{@{}c@{}}Easy\\(30 cm)\end{tabular}
        & \begin{tabular}[c]{@{}c@{}}Medium\\(50 cm)\end{tabular}
        & \begin{tabular}[c]{@{}c@{}}Hard\\(75 cm)\end{tabular}
        & \begin{tabular}[c]{@{}c@{}}OOD\\(80 cm)\end{tabular}
        & \begin{tabular}[c]{@{}c@{}}Easy\\(20 cm)\end{tabular}
        & \begin{tabular}[c]{@{}c@{}}Medium\\(30 cm)\end{tabular}
        & \begin{tabular}[c]{@{}c@{}}Hard\\(40 cm)\end{tabular}
        & \begin{tabular}[c]{@{}c@{}}OOD\\(43 cm)\end{tabular}
        & & \\
        \midrule
        Full CReF
        & \textbf{99.85}
        & \textbf{99.75}
        & \textbf{97.25}
        & \textbf{73.45}
        & \textbf{99.30}
        & \textbf{98.35}
        & \textbf{92.15}
        & \textbf{44.70}
        & \textbf{99.65}
        & \textbf{99.35}
        & \textbf{97.35}
        & \textbf{84.35}
        & 100(\textit{20}/20)
        & \textbf{90.45} \\

        w/o Cross-Attn
        & 96.25
        & 87.50
        & 78.70
        & 34.00
        & 97.10
        & 97.70
        & 83.00
        & 8.60
        & 99.40
        & 98.75
        & 96.90
        & 64.80
        & 95(\textit{19}/20)
        & 78.56 \\

        w/o GRF
        & 99.60
        & 93.40
        & 83.60
        & 49.10
        & 98.00
        & 98.25
        & 90.05
        & 29.95
        & 99.45
        & 99.15
        & 96.20
        & 68.55
        & 95(\textit{19}/20)
        & 83.78 \\

        w/o Highway Gate
        & 97.65
        & 98.40
        & 93.65
        & 50.20
        & 98.50
        & 96.90
        & 86.35
        & 17.20
        & 98.45
        & 97.65
        & 91.45
        & 73.10
        & 100(\textit{20}/20)
        & 83.29 \\

        HPL
        & 95.45
        & 81.45
        & 71.05
        & 27.40
        & 97.45
        & 94.40
        & 78.25
        & 20.85
        & 98.05
        & 92.65
        & 81.80
        & 55.15
        & 5(\textit{1}/20)
        & 74.57 \\
        \bottomrule
    \end{tabular}}
\end{table*}

\begin{table}[t]
    \caption{STAIR FAILURES IN ASCENT AND DESCENT.}
    \label{tab:stairs_failure_breakdown}
    \centering
    \scriptsize
    \setlength{\tabcolsep}{2.8pt}
    \renewcommand{\arraystretch}{1.08}
    \resizebox{\columnwidth}{!}{
    \begin{tabular}{lcccccccc}
        \toprule
        \multirow{2}{*}{Method} &
        \multicolumn{2}{c}{Easy} &
        \multicolumn{2}{c}{Medium} &
        \multicolumn{2}{c}{Hard} &
        \multicolumn{2}{c}{OOD} \\
        \cmidrule(lr){2-3}\cmidrule(lr){4-5}\cmidrule(lr){6-7}\cmidrule(lr){8-9}
        & Up & Down & Up & Down & Up & Down & Up & Down \\
        \midrule
        Full CReF        & 1   & \textbf{2}   & \textbf{2}   & \textbf{3}   & \textbf{3}   & \textbf{52}  & \textbf{93}  & \textbf{438} \\
        w/o Cross-Attn   & \textbf{0}   & 75  & 47  & 203 & 88  & 338 & 496 & 824 \\
        w/o GRF          & 2   & 6   & 25  & 107 & 39  & 289 & 128 & 890 \\
        w/o Highway Gate & 13  & 34  & 3   & 29  & 39  & 88  & 293 & 703 \\
        HPL              & 29  & 62  & 136 & 235 & 241 & 338 & 493 & 959 \\
        \bottomrule
    \end{tabular}}
\end{table}

\subsection{Simulation Experiments}
\label{subsec:sim_exp}

This subsection reports the simulation results of CReF. We focus on architectural comparison, stair foothold distribution analysis, and highway-gate behavior analysis.

\subsubsection{Architectural Comparison}
\label{subsubsec:arch_comp}

Table~\ref{tab:arch_comp} summarizes the architectural comparison. For the trained terrain categories, the stair entries are indexed by riser height and tread depth, while the gap and platform entries are indexed by gap width and platform height, all in centimeters. For stairs, gap, and platform, the OOD columns denote difficulty levels outside the training range. In the additional MuJoCo OOD evaluation, the values in parentheses denote the numbers of successful terrains out of 20 test terrains. The Overall score is computed from the stair, gap, and platform success rates only, excluding MuJoCo. Full CReF achieves the best overall performance and consistently outperforms the ablated variants and the perceptive baseline HPL, with larger advantages on harder and out-of-distribution terrains, including the additional MuJoCo evaluation. These results demonstrate improved robustness and generalization.

Among the ablations, removing cross-modal attention causes the largest degradation, especially on hard and out-of-distribution terrains, indicating that proprioception-conditioned depth querying is crucial for extracting terrain-scale cues for adaptive traversal. Without it, the policy tends to fall back to an average-terrain locomotion pattern, with less adaptive foot clearance and weaker generalization beyond the training range, which also explains the noticeable drop even on easy descending stairs. The particularly low success rate on the OOD gap terrain is mainly due to the evaluation protocol: large-gap traversal requires high actuation, but success rate is averaged over a range of commanded speeds, so low-command cases fail disproportionately and pull down the aggregate result. By comparison, removing GRF or the highway gate causes a milder but consistent decline, mainly on more challenging terrains, indicating that these modules improve the stability of multimodal fusion and temporal feature usage under demanding traversal conditions. For the HPL baseline, depth features are learned under privileged terrain guidance, making severe depth-distribution shifts more likely to generate unreliable geometry-related cues. CReF instead keeps control in a raw-depth feature space with recurrent temporal integration, which can help recent observations complement incomplete or irregular current depth frames.

Table~\ref{tab:stairs_failure_breakdown} further analyzes the failed stair trials from the same 2000 environment evaluation rollouts used for Table~\ref{tab:arch_comp}. Across all methods, descending failures increase much more rapidly than ascending failures as stair difficulty rises, indicating that stair descent is the more challenging regime. Full CReF consistently yields the fewest descending failures, especially on hard and OOD stairs, which is consistent with improved anticipatory foothold selection and temporal integration.

\begin{figure}[t]
  \centering
  \includegraphics[width=\linewidth]{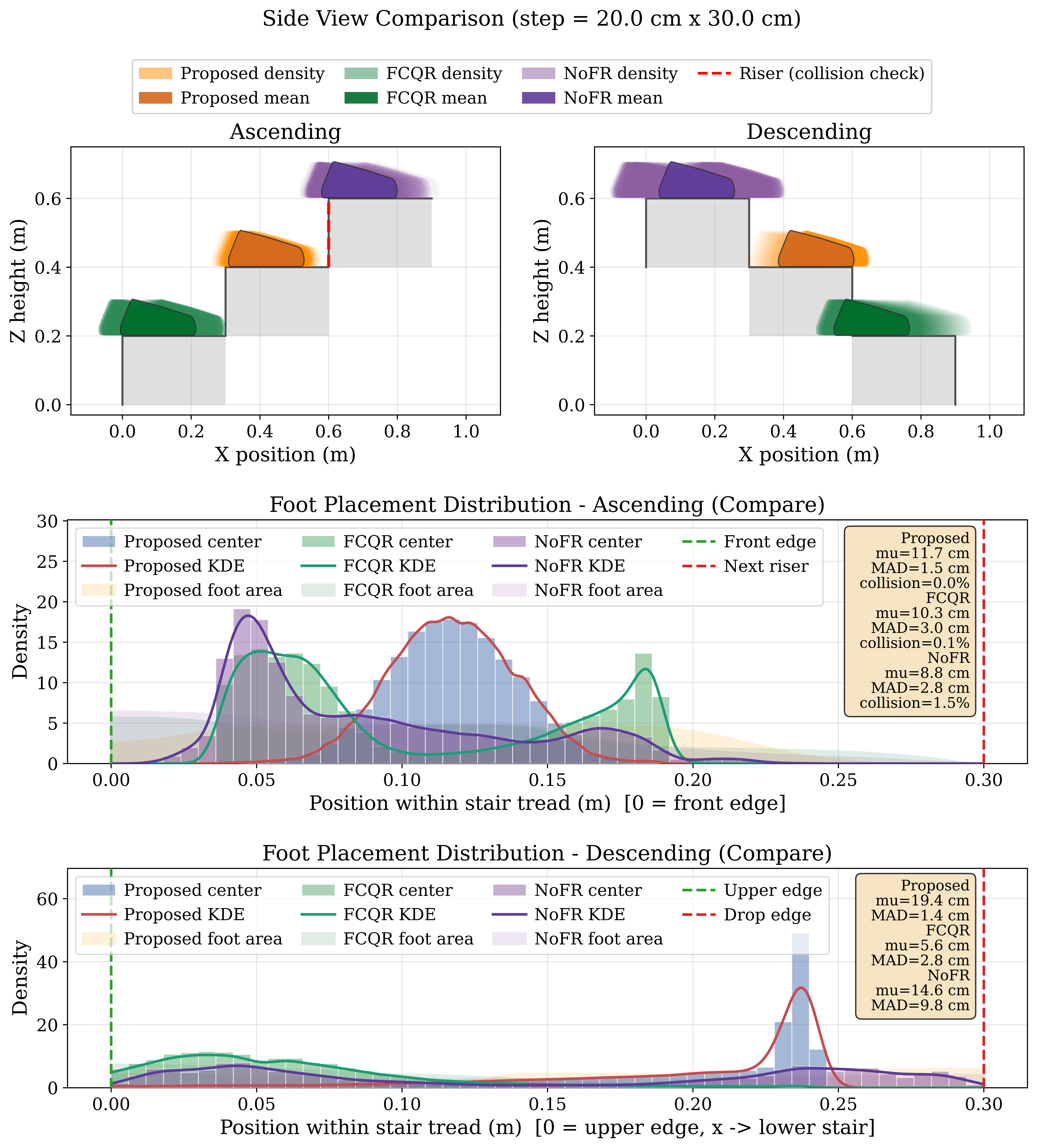}
  \caption{Stair foothold distribution comparison between the proposed foothold placement reward and the FCQR baseline. The proposed reward produces more concentrated and repeatable touchdown distributions in both ascent and descent, reduces touchdown deviation within the stair tread, and eliminates the logged ankle-riser collisions in ascending rollouts.}
  \label{fig:foot_placement_compare}
\end{figure}

\subsubsection{Stair Foothold Distribution Analysis}
\label{subsubsec:foothold_analysis}

Fig.~\ref{fig:foot_placement_compare} compares the stair touchdown distributions produced by the foothold placement reward, FCQR, and NoFR. The distributions are computed from converged rollouts on the hard stair setting, by pooling touchdown samples from both feet across all evaluated environments. The reported touchdown positions are the projected contact points of the foot-end geometric centers at touchdown. The proposed reward yields the tightest and most repeatable contact placement in both ascent and descent. On ascending stairs, it achieves the highest concentration and eliminates logged ankle-riser collisions. On descending stairs, it reduces the median absolute deviation from 2.8\,cm (FCQR) and 9.8\,cm (NoFR) to 1.4\,cm.

The improvement is not merely a shift of the touchdown mean. Instead, the entire distribution becomes more concentrated, indicating that the reward improves contact precision and repeatability. This is consistent with its design: although computed at touchdown, it provides an anticipatory shaping objective toward locally supportable regions, rather than only discouraging unfavorable contacts after they occur.

\begin{figure*}[t]
    \centering
    \vspace{7pt}
    \includegraphics[width=\textwidth]{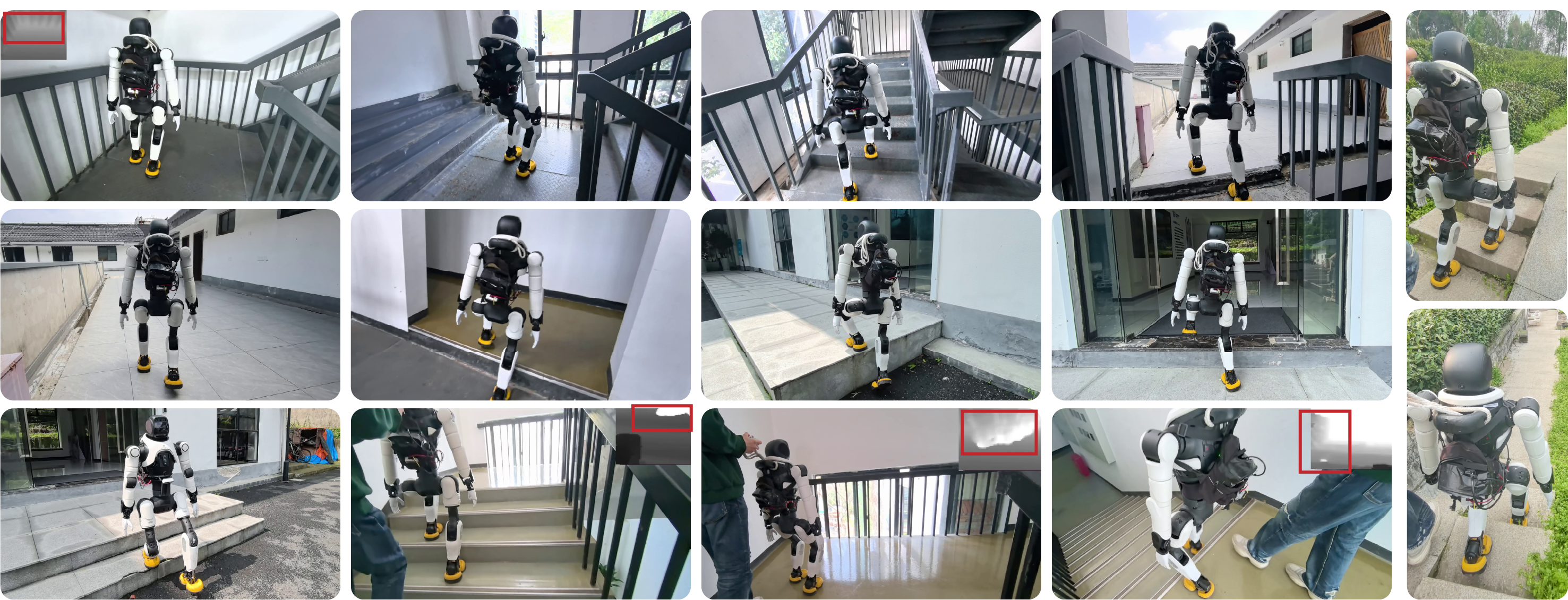}
    \caption{Representative real-world rollouts of CReF across multiple terrains and deployment scenes. The figure includes stair traversal with side railings, entrance-step and platform-like transitions, outdoor pathways, and other real-world terrain configurations. Red boxes highlight examples where environmental factors introduce out-of-distribution depth observations, such as large invalid regions in the sensed depth image.}
    \label{fig:real_world}
\end{figure*}

\begin{figure}[t]
    \centering
    \includegraphics[width=\linewidth]{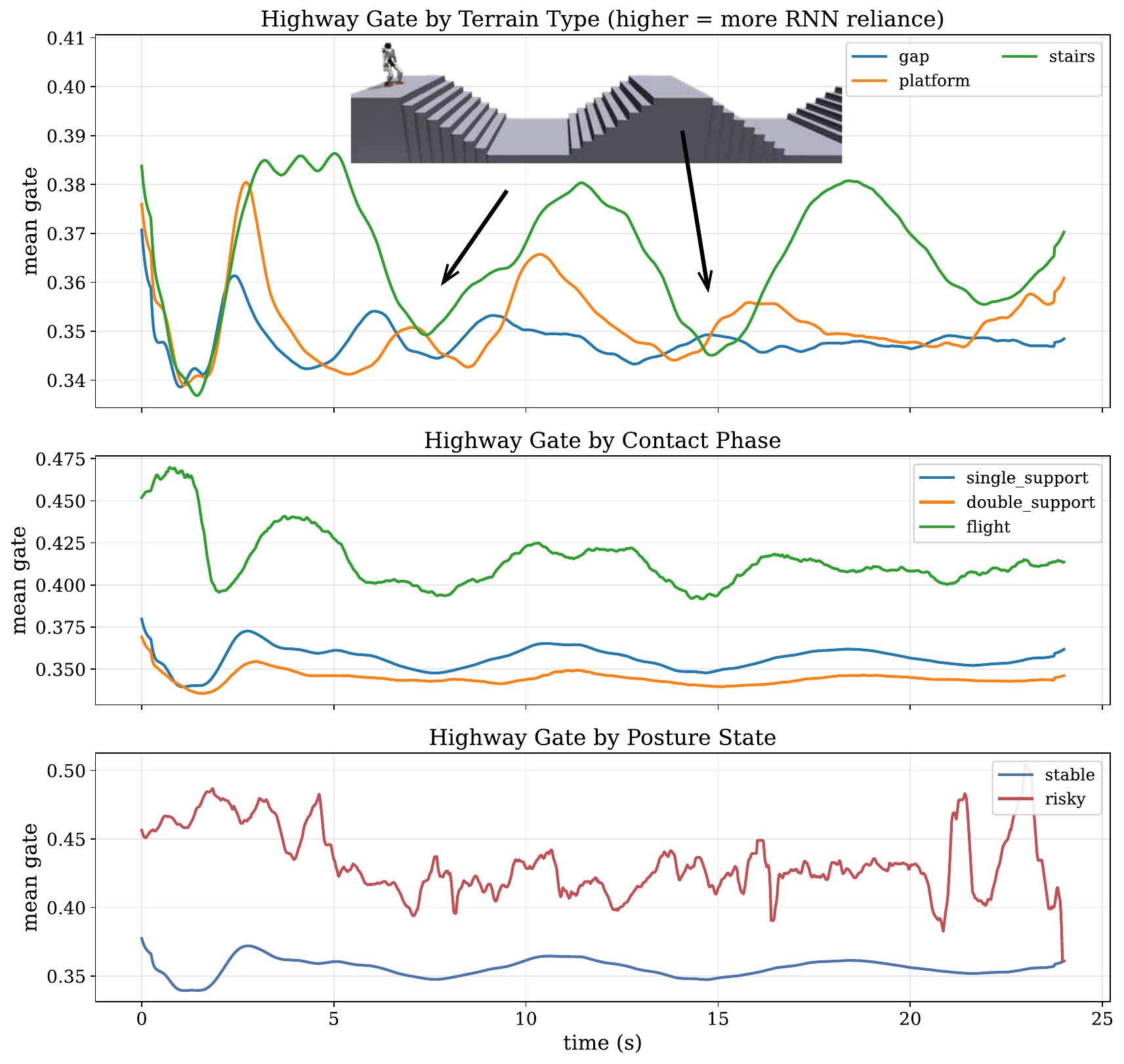}
    \caption{Highway gate behavior under different traversal conditions. Higher gate values indicate stronger reliance on recurrent features.}
    \label{fig:gate_behavior}
\end{figure}

\subsubsection{Highway Gate Behavior Analysis}
\label{subsubsec:gate_analysis}

Fig.~\ref{fig:gate_behavior} analyzes the learned highway gate, where larger values indicate stronger reliance on recurrent features. For analysis, gate values are averaged over channels and then aggregated within each state group. We define \emph{risky} states as timesteps satisfying $|\mathrm{roll}|>0.20$~rad or $|\mathrm{pitch}|>0.20$~rad, and treat the remaining timesteps as \emph{stable} states. Flight phases are identified from the foot-contact pattern. Across terrain types, the gate dynamically takes higher values on step-like terrains than on flat terrain over continuous rollout trajectories, showing that the policy recruits temporal memory more strongly when precise terrain anticipation is required. Across contact phases, gate values are higher during flight than during stable support. Across posture states, risky states consistently induce larger gate activation than stable states.

These trends are consistent with the intended function of the highway gate: recurrent features are emphasized when instantaneous observations are less sufficient for reliable control, and attenuated when feedforward perception is adequate.

\begin{table}[t]
    \caption{INDOOR REAL-WORLD TRAVERSAL RESULTS.}
    \label{tab:real_world_sr}
    \centering
    \scriptsize
    \setlength{\tabcolsep}{4.2pt}
    \renewcommand{\arraystretch}{1.10}
    \begin{tabular}{lcc}
        \toprule
        Task & Setting & Success / Trials \\
        \midrule
        Stairs & Indoor, 15\,cm / 30\,cm & 20/20 \\
        Platform & Indoor, 40\,cm & 20/20 \\
        Gap & Indoor, 80\,cm & 18/20 \\
        OOD terrain & Indoor OOD scene & 19/20 \\
        \bottomrule
    \end{tabular}
\end{table}

\subsection{Real-World Experiments}
\label{subsec:real_world}

Real-world experiments validate zero-shot transfer of CReF to a humanoid robot across stair traversal, platform ascent, gap crossing, and diverse indoor and outdoor scenes. The indoor OOD scene follows the MuJoCo OOD setting and includes hollow pallet-like structures and side clutter, as shown in the middle row of Fig.~\ref{fig:cover}. Table~\ref{tab:real_world_sr} reports representative indoor quantitative evaluations, where each task is assessed over 20 trials. For stairs, one trial consists of a complete ascent-descent traversal on indoor stairs with 15\,cm rise and 30\,cm tread. On the AGIBOT X2 Ultra, CReF achieves strong performance across these standardized indoor tasks, including successful traversal of a 40\,cm platform and an 80\,cm gap. Beyond these evaluations, CReF further demonstrates stable deployment in broader real-world environments.

Fig.~\ref{fig:real_world} shows representative real-world rollouts. The policy stably climbs real stairs with side railings at a rise of 20\,cm and a tread of 26\,cm, even under changing illumination. It also maintains stable locomotion under severe depth degradation, including large invalid depth holes caused by strong reflection, and in outdoor scenes with dense vegetation, asymmetric boundaries, and other non-traversable structures that induce substantial depth out-of-distribution effects. These results suggest that CReF can still exploit useful cues under degraded visual input, although such artifacts may still affect traversal behavior.

\section{Conclusions}

This paper presents CReF, which maps proprioception and forward-facing depth directly to joint position targets without explicit geometric intermediates. By combining cross-modal attention, gated residual fusion, recurrent fusion, and a terrain-aware foothold placement reward, CReF improves terrain anticipation and foot placement consistency, with clear gains on stair traversal. Experiments in simulation and on hardware demonstrate strong terrain-traversal performance and zero-shot transfer to challenging real-world scenes.

At the same time, certain limitations still exist. Depth alone does not preserve appearance and texture cues that may improve perceptual robustness. A natural direction for future work is therefore to investigate binocular RGB-based sensing, which can jointly provide depth structure and texture information for richer scene representations and stronger adaptability across diverse real-world environments.

\bibliographystyle{IEEEtran}

\bibliography{references}

\end{document}